\documentclass[11pt,a4paper]{article}
\usepackage[hyperref]{acl2020}
\usepackage{times}
\usepackage{latexsym}
\usepackage{amsmath}
\usepackage{amssymb}
\usepackage{microtype}
\usepackage{graphicx}
\usepackage{multirow}
\usepackage{tabularx}
\usepackage{subcaption}
\usepackage{xcolor}
\usepackage{adjustbox}

\usepackage{censor}

\aclfinalcopy

\title{Joint Modelling of Emotion and Abusive Language Detection}

\author{Santhosh Rajamanickam\\
  ILLC, University of Amsterdam\\
  \texttt{rajamanickamsanthosh@gmail.com} \\\And
  Pushkar Mishra\\
  Facebook AI\\
  \texttt{pushkarmishra@fb.com} \\\AND
  Helen Yannakoudakis \\
  Dept.of Informatics, King's College London\\
  \texttt{helen.yannakoudakis@kcl.ac.uk} \\\And
  Ekaterina Shutova \\
  ILLC, University of Amsterdam \\
  \texttt{e.shutova@uva.nl}}

\date{}

\begin{document}
\maketitle

\begin{abstract}
The rise of online communication platforms has been accompanied by some undesirable effects, such as the proliferation of aggressive and abusive behaviour online. Aiming to tackle this problem, the natural language processing (NLP) community has experimented with a range of techniques for abuse detection. While achieving substantial success, these methods have so far only focused on modelling the linguistic properties of the comments and the online communities of users, disregarding the emotional state of the users and how this might affect their language. The latter is, however, inextricably linked to abusive behaviour. In this paper, we present the first joint model of emotion and abusive language detection, experimenting in a multi-task learning framework that allows one task to inform the other. Our results demonstrate that incorporating affective features leads to significant improvements in abuse detection performance across datasets.
\end{abstract}

\section{Introduction}
Aggressive and abusive behaviour online can lead to severe psychological consequences for its victims \cite{munro2011}. This stresses the need for automated techniques for abusive language detection, a problem that has recently gained a great deal of interest in the natural language processing community. The term \textit{abuse} refers collectively to all forms of expression that vilify or offend an individual or a group, including \textit{racism, sexism, personal attacks, harassment, cyber-bullying}, and many others. Much of the recent research has focused on detecting \textit{explicit} abuse, that comes in the form of expletives, derogatory words or threats, with substantial success \cite{Mishra2019Survey}. However, abuse can also be expressed in more implicit and subtle ways, for instance, through the use of ambiguous terms and figurative language, which has proved more challenging to identify.

The NLP community has experimented with a range of techniques for abuse detection, such as recurrent and convolutional neural networks \cite{pavlopoulos,park2017one,wang2018interpreting}, character-based models \cite{nobata} and graph-based learning methods \cite{mishra2018author, aglionby, mishra-gcn}, obtaining promising results. However, all of the existing approaches have focused on modelling the linguistic properties of the comments or the meta-data about the users.
On the other hand, abusive language and behaviour are also inextricably linked to the emotional and psychological state of the speaker \cite{patrick1901psychology}, which is reflected in the affective characteristics of their language \cite{mabry1974dimensions}. In this paper, we propose to model these two phenomena jointly and present the first abusive language detection method that incorporates affective features via a multitask learning (MTL) paradigm.

MTL \citep{caruana1997multitask} allows two or more tasks to be learned jointly, thus sharing information and features between the tasks. In this paper, our main focus is on abuse detection; hence we refer to it as the \textit{primary task}, while the task that is used to provide additional knowledge --- emotion detection --- is referred to as the \textit{auxiliary task}. We propose an MTL framework where a single model can be trained to perform emotion detection and identify abuse at the same time. We expect that affective features, which result from a joint learning setup through shared parameters, will encompass the emotional content of a comment that is likely to be predictive of potential abuse. 

We propose and evaluate different MTL architectures. We first experiment with hard parameter sharing, where the same encoder is shared between the tasks. We then introduce two variants of the MTL model to relax the hard sharing constraint and further facilitate positive transfer. Our results demonstrate that the MTL models significantly outperform single-task learning (STL) in two different abuse detection datasets. This confirms our hypothesis of the importance of affective features for abuse detection. 
Furthermore, we compare the performance of MTL to a transfer learning baseline and demonstrate that MTL provides significant improvements over transfer learning. 

\section{Related Work}
Techniques for abuse detection have gone through several stages of development, starting with extensive manual feature engineering and then turning to deep learning. Early approaches experimented with lexicon-based features \cite{gitari2015lexicon}, bag-of-words (\textit{BOW}) or \textit{n-gram} features \cite{sood2012using,dinakar2011modeling}, and user-specific features, such as age \citep{dadvar2013improving} and gender \citep{waseem2016hateful}.

With the advent of deep learning, the trend shifted, with abundant work focusing on neural architectures for abuse detection. In particular, the use of convolutional neural networks (\textit{CNN}s) for detecting abuse has shown promising results \cite{park2017one,wang2018interpreting}. This can be attributed to the fact that \textit{CNN}s are well suited to extract local and position-invariant features \cite{yin2017comparative}. Character-level features have also been shown to be beneficial in tackling the issue of Out-of-Vocabulary (OOV) words \citep{mishra2018neural}, since abusive comments tend to contain obfuscated words. 
Recently, approaches to abuse detection have moved towards more complex models that utilize auxiliary knowledge in addition to the abuse-annotated data. For instance, \citet{mishra2018author, mishra-gcn} used community-based author information as features in their classifiers with promising results. \citet{founta2019unified} used transfer learning to fine-tune features from the author metadata network to improve abuse detection.

MTL, introduced by \citet{caruana1997multitask}, has proven successful in many NLP problems, as illustrated in the MTL survey of \citet{zhang2017survey}. It is interesting to note that many of these problems are domain-independent tasks, such as part-of-speech tagging, chunking, named entity recognition, etc. \cite{collobert2008unified}. These tasks are not restricted to a particular dataset or domain, i.e., any text data can be annotated for the phenomena involved. On the contrary, tasks such as abuse detection are domain-specific and restricted to a handful of datasets (typically focusing on online communication), therefore presenting a different challenge to MTL. 

Much research on emotion detection cast the problem in a categorical framework, identifying specific classes of emotions and using e.g., Ekman's model of six emotions \citep{ekman1992argument}, namely anger, disgust, fear, happiness, sadness, surprise. Other approaches adopt the Valence-Arousal-Dominance (\textit{VAD}) model of emotion \cite{mehrabian1996pleasure}, which represents polarity, degree of excitement, and degree of control, each taking a value from a range. The community has experimented with a variety of computational techniques for emotion detection, including vector space modelling \cite{danisman2008feeler}, machine learning classifiers \cite{perikos2016recognizing} and deep learning methods \cite{zhang2018text}. In their work, \citet{zhang2018text} take an MTL approach to emotion detection. However, all the tasks they consider are emotion-related (annotated for either classification or emotion distribution prediction), and the results show improvements over single-task baselines. \citet{akhtar2018multi} use a multitask ensemble architecture to learn emotion, sentiment, and intensity prediction jointly and show that these tasks benefit each other, leading to improvements in performance. To the best of our knowledge, there has not yet been an approach investigating emotion in the context of abuse detection.

\section{Datasets}
The tasks in an MTL framework should be related in order to obtain positive transfer. MTL models are sensitive to differences in the domain and distribution of data \cite{pan2009survey}. This affects the stability of training, which may deteriorate performance in comparison to an STL model \cite{zhang2017survey}. 
We experiment with abuse and emotion detection datasets\footnote{We do not own any rights to the datasets (or the containing tweets). In the event of one who wishes to attain any of the datasets, to avoid redistribution infringement, we request them to contact the authors/owners of the source of the datasets.} that are from the same data domain --- Twitter. All of the datasets were subjected to the same pre-processing steps, namely lower-casing, mapping all \textit{mentions} and \textit{URLs} to a common token (i.e., \textit{\_MTN\_} and \textit{\_URL\_}) and mapping hashtags to words. 

\subsection{Abuse detection task}
To ensure that the results are generalizable, we experiment with two different abuse detection datasets. 

\paragraph{OffensEval 2019 (\textit{OffensEval})} This dataset is from \textit{SemEval 2019 - Task 6: OffensEval 2019 - Identifying and Categorizing Offensive Language in Social Media} \cite{zampieri2019predicting,zampieri2019semeval}. We focus on Subtask A, which involves offensive language identification. 
It contains $13,240$ annotated tweets, and each tweet is classified as to whether it is offensive ($33\%$) or not ($67\%$). Those classified as offensive contain offensive language or targeted offense, which includes insults, threats, profane language and swear words. The dataset was annotated using crowdsourcing, with gold labels assigned based on the agreement of three annotators. 

\paragraph{Waseem and Hovy 2016 (\textit{Waseem\&Hovy})} This dataset was compiled by  \citet{waseem2016hateful} by searching for commonly used slurs and expletives related to religious, sexual, gender and ethnic minorities. The tweets were then annotated with one of three classes: \textit{racism}, \textit{sexism} or \textit{neither}. The annotations were subsequently checked through an expert review, which yielded an inter-annotator agreement of $\kappa=0.84$. The dataset contains $16,907$ TweetIDs and their corresponding annotation, out of which only $16,202$ TweetIDs were retrieved due to users being reported or tweets having been taken down since it was first published in 2016. The distribution of classes is: $1,939$ ($12\%$) \textit{racism}; $3,148$ ($19.4\%$) \textit{sexism}; and $11,115$ ($68.6\%$) \textit{neither},  which is comparable to the original distribution: ($11.7\%$ : $20.0\%$ : $68.3\%$).\\

It should be noted that racial or cultural biases may arise from annotating data using crowdsourcing, as pointed out by \citet{sap2019risk}. The performance of the model depends on the data used for training, which in turn depends on the quality of the annotations and the experience level of the annotators. However, the aim of our work is to investigate the relationship between emotion and abuse detection, which is likely to be independent of the biases that may exist in the annotations.  

\subsection{Emotion detection task}
\paragraph{Emotion (\textit{SemEval18})} This dataset is from \textit{SemEval-2018 Task 1: Affect in Tweets} \cite{mohammad2018semeval}, and specifically from Subtask 5 which is a multilabel classification of $11$ emotion labels that best represent the mental state of the author of a tweet. The dataset consists of around $11$k tweets (training set: $6839$; development set: $887$; test set: $3260$). It contains the TweetID and $11$ emotion labels (\textit{anger}, \textit{anticipation}, \textit{disgust}, \textit{fear}, \textit{joy}, \textit{love}, \textit{optimism}, \textit{pessimism}, \textit{sadness}, \textit{surprise}, \textit{trust}) which take a binary value to indicate the presence or absence of the emotion. The annotations were obtained for each tweet from at least $7$ annotators and aggregated based on their agreement. 

\section{Approach}
In this section, we describe our baseline models and then proceed by describing our proposed models for jointly learning to detect emotion and abuse. 

\subsection{Single-Task Learning}
As our baselines, we use different Single-Task Learning (STL) models that utilize abuse detection as the sole optimization objective. The STL experiments are conducted for each primary-task dataset separately. Each STL model takes as input a sequence of words $\{w_{\textrm{1}}, w_{\textrm{2}}, ..., w_{\textrm{n}}\}$, which are initialized with $k$-dimensional vectors $\textrm{e}$ from a pre-trained embedding space. We experiment with two different architecture variants:  

\paragraph{Max Pooling and MLP classifier} We refer to this baseline as STL$_{maxpool+MLP}$. In this baseline, a two-layered bidirectional Long Short-Term Memory (LSTM) network \cite{hochreiter1997long} is applied to the embedding representations $\textrm{e}$ of words in a post to get contextualized word representations $\{\textrm{h}_{\textrm{1}}, \textrm{h}_{\textrm{2}}, ..., \textrm{h}_{\textrm{n}}\}$:

\begin{equation}
\textrm{h}_{\textrm{t}} = [\overrightarrow{\textrm{h}_{\textrm{t}}};\overleftarrow{\textrm{h}_{\textrm{t}}}]
\end{equation}

\noindent with $\overrightarrow{\textrm{h}_{\textrm{t}}}, \overleftarrow{\textrm{h}_{\textrm{t}}}\in \mathbb{R}^{l}$ and $\textrm{h}_{\textrm{t}} \in \mathbb{R}^{2\cdot l}$, where $l$ is the hidden dimensionality of the BiLSTM. We then apply a max pooling operation over $\{\textrm{h}_{\textrm{1}}, \textrm{h}_{\textrm{2}}, ..., \textrm{h}_{\textrm{n}}\}$: 

\begin{equation}
  r^{\textrm{(p)}}_{\textrm{i}} = {max_\textrm{i}}(\textrm{h}_{\textrm{1}}, \textrm{h}_{\textrm{2}}, ..., \textrm{h}_{\textrm{n}}) 
\end{equation}

\noindent where $\textrm{r}^\textrm{(p)}\in \mathbb{R}^{2\cdot l}$ and where the superscript $\textrm{(p)}$ is used to indicate that the representations correspond to the primary task. This is followed by dropout \cite{srivastava2014dropout} for regularization and a $2$-layered Multi-layer Perceptron (MLP) \cite{hinton1987learning}:

\begin{align}
    \textrm{m}^\textrm{1(p)} &= BatchNorm(tanh(W^{l_1}{\textrm{r}^{\textrm{(p)}}})) \\ \label{eq:mlp}
  \textrm{m}^\textrm{2(p)} &= tanh(W^{l_2}{\textrm{m}}^\textrm{1(p)}) \\
    \textrm{m}_{\textrm{t}}^\textrm{(p)} &= \textrm{m}_{\textrm{t}}^\textrm{2(p)} \label{eq:mlp_end}
\end{align}

\noindent where $W^{l_1}$ and $W^{l_2}$ are the weight matrices of the $2$-layer MLP. Dropout is applied to the output $\textrm{m}^\textrm{(p)}$ of the MLP, which is then followed by a linear output layer to get the unnormalized output $\textrm{o}^\textrm{(p)}$. For \textit{OffensEval}, a sigmoid activation $\sigma$ is then applied in order to make a binary prediction with respect to whether a post is offensive or not, while the network parameters are optimized to minimize the binary cross-entropy (BCE):
\begin{multline}
    L_{BCE}=-\frac{1}{N}\sum^N_{i=1}y_i\cdot log(p(y_i)) + \\ 
    (1-y_i)\cdot log(1-p(y_i)) 
    \label{eq:bce}
\end{multline}

\noindent where $N$ is the number of training examples, and $y$ denotes the true and $p(y)$ the predicted label. For \textit{Waseem\&Hovy}, a $log\_softmax$ activation is applied for multiclass classification, while the network parameters are optimized to minimize the categorical cross-entropy, that is, the negative log-likelihood (NLL) of the true labels: 

\begin{equation}
    L_{NLL}=-\frac{1}{N}\sum^N_{i=1}log(p(y_i))  
    \label{eq:nll}
\end{equation}

\begin{figure*}[!htb]
\centering
 \resizebox{0.65\textwidth}{!}{
	\includegraphics{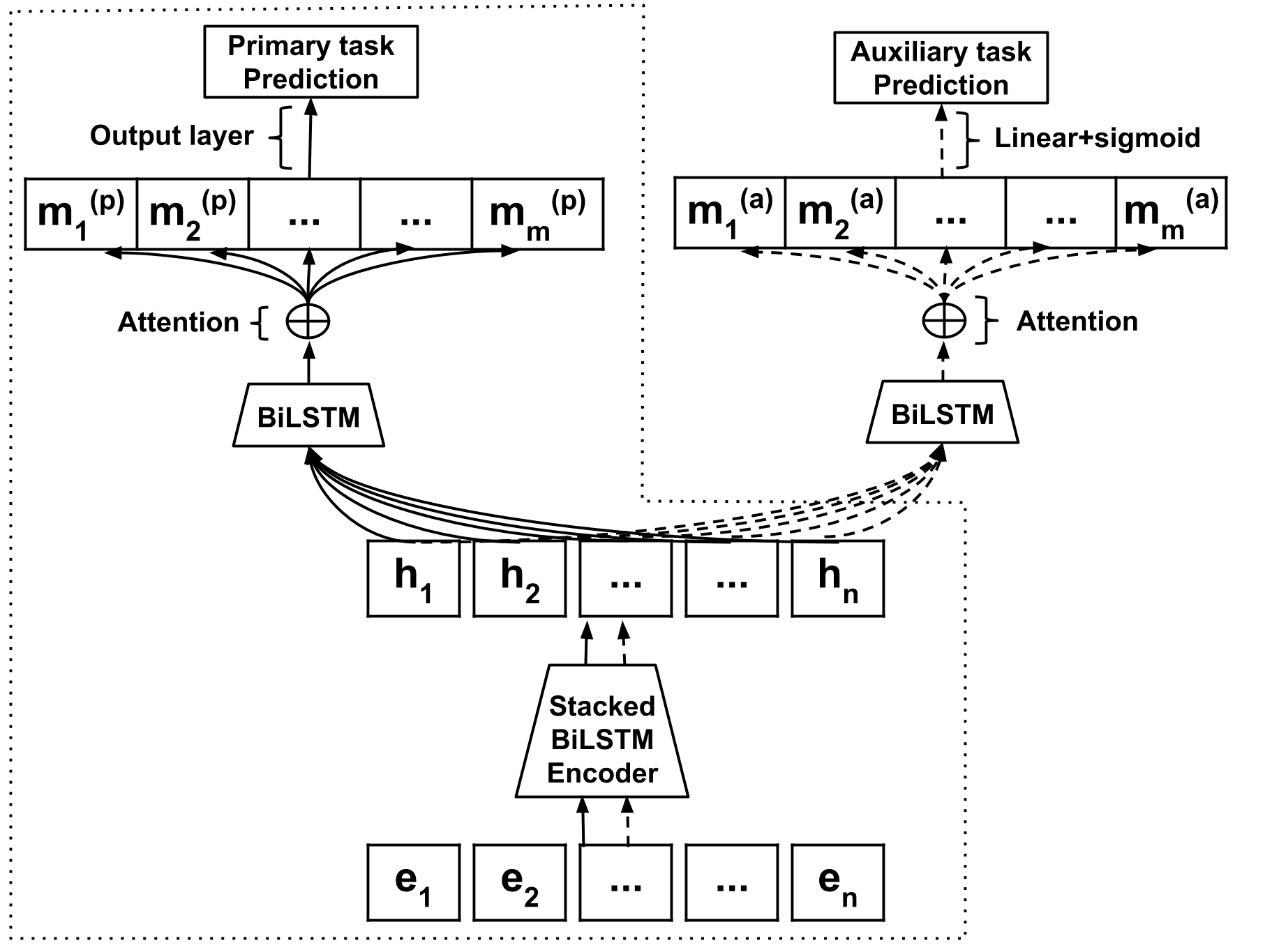}}
	\caption{MTL \textit{Hard Sharing} model. The embedding representations $\{\textrm{e}_{\textrm{1}}, \textrm{e}_{\textrm{2}}, ..., \textrm{e}_{\textrm{n}}\}$ are either a result of projection through the GloVe embedding layer or a concatenation of the projections through the GloVe and ELMo embedding layer. The different arrows are used to indicate the different passes for the primary and auxiliary task. The units on the left-hand side correspond to the primary task and the units on the right-hand side correspond to the auxiliary task with the \textit{Stacked BiLSTM Encoder} and embedding layers shared by both tasks. The model inside the dotted box corresponds to the  STL$_{BiLSTM+attn}$ architecture.} 
	\label{fig:vanilla_hard}
\end{figure*}

\paragraph{BiLSTM and Attention classifier} 
We refer to this model as STL$_{BiLSTM+attn}$. In this baseline (Figure \ref{fig:vanilla_hard}; enclosed in the dotted boxes), rather than applying max pooling, we apply dropout to $\textrm{h}$ which is then followed by a third BiLSTM layer and an attention mechanism:

\begin{align}
    \textrm{u}_{\textrm{t}}^\textrm{(p)} &= W^a\textrm{r}_{\textrm{t}}^\textrm{(p)} \label{eq:attention_start} \\ 
    a_{\textrm{t}}^\textrm{(p)} &= \frac{exp(\textrm{u}_{\textrm{t}}^\textrm{(p)})}{\sum_{\textrm{t}} exp(\textrm{u}_{\textrm{t}}^\textrm{(p)})} \\
    \textrm{m}^\textrm{(p)} &= \sum_{\textrm{t}} a_{\textrm{t}}^\textrm{(p)} \textrm{r}_{\textrm{t}}^\textrm{(p)} \label{eq:attention_end}
\end{align}

\noindent where $\textrm{r}^\textrm{(p)}$ is the output of the third BiLSTM. We then apply dropout to the output of the attention layer $\textrm{m}^\textrm{(p)}$. The remaining components, output layer and activation, are the same as the STL$_{maxpool+MLP}$ model. \\

Across the two STL baselines, we further experiment with two different input representations: 1) GloVe (G), where the input is projected through the GloVe embedding layer \cite{pennington2014glove}; 
2) GloVe+ELMo (G+E), where the input is first projected through the GloVe embedding layer and the ELMo embedding layer \cite{peters2018deep} separately, and then the final word representation $\textrm{e}$ is obtained by concatenating the output of these two layers. Given these input representations, we have a total of $4$ different baseline models for abuse detection. We use grid search to tune the hyperparameters of the baselines on the development sets of the primary task (i.e., abuse detection). 

\begin{figure*}[!htb]
\centering
\resizebox{0.75\textwidth}{!}{
	\includegraphics{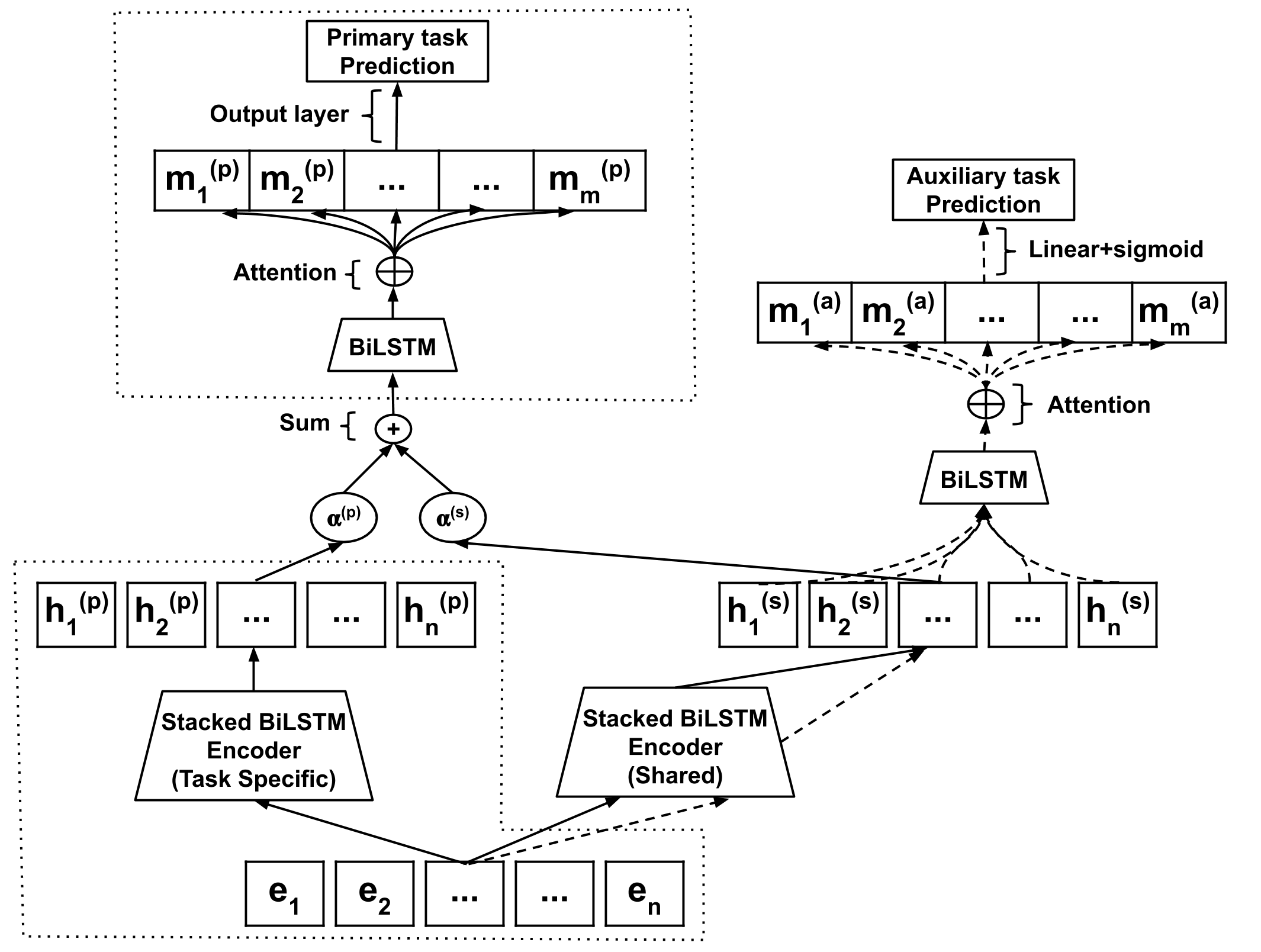}}
	\caption{MTL \textit{(Gated) Double Encoder} architecture. For the MTL Gated Double Encoder model we use two learnable parameters $\alpha$ that control information flow. For the MTL \textit{Double Encoder} model, these are fixed and set to $1$. The dotted boxes represent the STL$_{BiLSTM+attn}$ architecture.}
	\label{fig:detach_soft_encoder}
\end{figure*}

\subsection{Multi-task Learning}
Our MTL approach uses two different optimization objectives: one for abuse detection and another for emotion detection. The two objectives are weighted by a hyperparameter $\beta$ [($1-\beta$) for abuse detection and $\beta$ for emotion detection] that controls the importance we place on each task. 
We experiment with different STL architectures for the auxiliary task and propose MTL models that contain two network branches -- one for the primary task and one for the auxiliary task -- connected by a shared encoder which is updated by both tasks alternately. 

\paragraph{Hard Sharing Model}
This model architecture, referred to as MTL$_{Hard}$, is inspired by \citet{caruana1997multitask} and uses \textit{hard parameter sharing}: it consists of a single encoder that is shared and updated by both tasks, followed by task-specific branches. Figure \ref{fig:vanilla_hard} presents MTL$_{Hard}$ where the dotted box represents the STL$_{BiLSTM+attn}$ architecture that is specific to the abuse detection task. In the right-hand side branch -- corresponding to the auxiliary objective of detecting emotion -- we apply dropout to $\textrm{h}$ before passing it to a third BiLSTM. This is then followed by an attention mechanism to obtain $\textrm{m}^\textrm{(a)}$ and then dropout is applied to it. The superscript $\textrm{(a)}$ is used to indicate that these representations correspond to the auxiliary task. Then, we obtain the unnormalized output $\textrm{o}^{\textrm{(a)}}$ after passing $\textrm{m}^\textrm{(a)}$ through a linear output layer with $\textrm{o}^{\textrm{(a)}}\in \mathbb{R}^{11}$ ($11$ different emotions in \textit{SemEval18}), which is then subjected to a sigmoid activation to obtain a prediction $p(y)$. While the primary task on the left is optimized using either Equation \ref{eq:bce} or \ref{eq:nll} (depending on the dataset used), the auxiliary task is optimized to minimize binary cross-entropy. 

\paragraph{Double Encoder Model }
This model architecture, referred to as MTL$_{DEncoder}$, is an extension of the previous model that now has two BiLSTM encoders: a task-specific two-layered BiLSTM encoder for the primary task, and a shared two-layered BiLSTM encoder. During each training step of the primary task, the input representation $\textrm{e}$ for the primary task is passed through both encoders, which results in two contextualized word representations $\{\textrm{h}_{\textrm{1}}^{\textrm{(p)}}, \textrm{h}_{\textrm{2}}^{\textrm{(p)}}, ..., \textrm{h}_{\textrm{n}}^{\textrm{(p)}}\}$ and $\{\textrm{h}_{\textrm{1}}^{\textrm{(s)}}, \textrm{h}_{\textrm{2}}^{\textrm{(s)}}, ..., \textrm{h}_{\textrm{n}}^{\textrm{(s)}}\}$, where superscript (s) is used to denote the representations that result from the shared encoder. These are then summed (Figure \ref{fig:detach_soft_encoder}, where both $\alpha^{\textrm{(p)}}$ and $\alpha^{\textrm{(s)}}$ are fixed and set to $1$) and the output representation is passed through a third BiLSTM followed by an attention mechanism to get the post representation $\textrm{m}^{\textrm{(p)}}$. The rest of the components of the primary task branch, as well as the auxiliary task branch are the same as those in MTL$_{Hard}$. 

\paragraph{Gated Double Encoder Model } 
This model architecture, referred to as MTL$_{Gated DEncoder}$, is an extension of MTL$_{DEncoder}$, but is different in the way we obtain the post representations $\textrm{m}^{\textrm{(p)}}$. Representations $\textrm{h}^{\textrm{(p)}}$ and $\textrm{h}^{\textrm{(s)}}$ are now merged using two learnable parameters $\alpha^{\textrm{(p)}}$ and $\alpha^{\textrm{(s)}}$ (where $\alpha^{\textrm{(p)}}+\alpha^{\textrm{(s)}}=1.0$) to control the flow of information from the representations that result from the two encoders (Figure \ref{fig:detach_soft_encoder}): 
\begin{equation}
    \alpha^{\textrm{(p)}}\cdot \textrm{h}^{\textrm{(p)}} + \alpha^{\textrm{(s)}}\cdot \textrm{h}^{\textrm{(s)}}
\end{equation}
The remaining architecture components of the primary task and auxiliary task branch are the same as for MTL$_{DEncoder}$.

\section{Experiments and results}

\subsection{Experimental setup}

\paragraph{Hyperparameters} We use pre-trained GloVe embeddings\footnote{https://nlp.stanford.edu/projects/glove/} with dimensionality $300$ and pre-trained ELMo embeddings\footnote{https://allennlp.org/elmo} with dimensionality $1024$. Grid search is performed to determine the optimal hyperparameters. We find an optimal value of $\beta=0.1$ that makes the updates for the auxiliary task $10$ times less important. The encoders consist of $2$ stacked BiLSTMs with $hidden\_size=512$. For all primary task datasets, the BiLSTM+Attention classifier and the $2$-layered MLP classifier have $hidden\_size=256$. For the auxiliary task datasets, the BiLSTM+Attention classifier and the 2-layered MLP classifier have $hidden\_size=512$. Dropout is set to $0.2$. We use the \textit{Adam optimizer} \cite{kingma2014adam} for all experiments. All model weights are initialized using \textit{Xavier Initialization} \cite{glorot2010understanding}. For MTL$_{Gated DEncoder}$, $\alpha^{\textrm{(p)}}=0.9$ and $\alpha^{\textrm{(s)}}=0.1$.
 
 \vspace{-0.2cm}
\paragraph{Training} All models are trained until convergence for both the primary and the auxiliary task, and early stopping is applied based on the performance on the validation set. For MTL, we ensure that both the primary and the auxiliary task have completed at least $5$ epochs of training. The MTL training process involves randomly (with $p=0.5$) alternating between the abuse detection and emotion detection training steps. Each task has its own loss function, and in each of the corresponding task's training step, the model is optimized accordingly. 
All experiments are run using stratified $10$-fold cross-validation, and we use the paired t-test for significance testing. We evaluate the models using Precision ($P$), Recall ($R$), and F1 ($F1$), and report the average $macro$ scores across the $10$ folds. 

\subsection{STL experiments}
The STL experiments are conducted on the abuse detection datasets independently. As mentioned in the STL section, we experiment with four different model configurations to select the best STL baseline.

Table \ref{tab:baseline_offenseval} presents the evaluation results of the STL models trained and tested on the \textit{OffensEval} dataset, and Table \ref{tab:baseline_waseem} on the \textit{Waseem and Hovy} dataset. The best results are highlighted in bold and are in line with the validation set results. We select the best performing STL model configuration on each dataset and use it as part of the corresponding MTL architecture in the MTL experiments below.

\begin{table}[tb]
\centering
\resizebox{0.46\textwidth}{!}{
\begin{subtable}[t]{0.48\textwidth}
\begin{tabular}{|l|l|l|l|l|}\hline
\multicolumn{2}{|l|}{\textbf{STL model}} & \textbf{P} & \textbf{R} & \textbf{F1} \\ \hline

\multirow{3}{*}{G}  & \textit{maxpool+MLP}  & 76.35 & \textbf{73.34} & 74.24  \\  
                        & \textit{BiLSTM+attn}  & 77.34 & 72.77 & 73.97 \\ \hline
\multirow{3}{*}{G+E}   & \textit{maxpool + MLP}  & 77.19 & 72.73 & 73.95 \\  
                        & \textbf{\textit{BiLSTM+attn}}  & \textbf{77.40} & {73.27} & \textbf{74.40} \\ \hline
\end{tabular}
\caption{{\normalsize \textit{Twitter - OffensEval} STL results.}}
\label{tab:baseline_offenseval}
\end{subtable}
}
\bigskip

\resizebox{0.46\textwidth}{!}{
\begin{subtable}[t]{0.48\textwidth}
\begin{tabular}{|l|l|l|l|l|} \hline
\multicolumn{2}{|l|}{\textbf{STL model}} & \textbf{P} & \textbf{R} & \textbf{F1} \\ \hline

\multirow{3}{*}{G}  & \textbf{\textit{maxpool+MLP}}  & {79.39} & \textbf{78.20} & \textbf{78.33}  \\  
                        & \textit{BiLSTM+attn}  & 77.97 & 77.57 & 77.49  \\ \hline
\multirow{3}{*}{G+E}   & \textit{maxpool+MLP}  & \textbf{80.66} & 77.13 & 78.31 \\ 
                        & \textit{BiLSTM+attn}  & 79.08 & 77.93 & 78.16  \\ \hline
\end{tabular}
\caption{{\normalsize \textit{Twitter - Waseem and Hovy} STL results.}}
\label{tab:baseline_waseem}
\end{subtable}
}

\caption{STL model comparisons. In these tables, G denotes models that use GloVe embeddings and G+E denotes models in which word representations are concatenations of their corresponding GloVe and ELMo embeddings. The best performing model is highlighted in bold. }
\label{tab:baseline}
\end{table}

\subsection{MTL experiments}

In this section, we examine the effectiveness of the MTL models for the abuse detection task and explore the impact of using emotion detection as an auxiliary task. We also compare the performance of our MTL models with that of a transfer learning approach.

\paragraph{Emotion detection as an auxiliary task}
In this experiment, we test whether incorporating emotion detection as an auxiliary task improves the performance of abuse detection. Tables \ref{tab:mtl_offenseval_emotion} and  \ref{tab:mtl_waseem_emotion} show the results on \textit{OffensEval} and \textit{Waseem and Hovy} datasets ($\dagger$ indicates statistically significant results over the corresponding STL model). Learning emotion and abuse detection jointly proved beneficial, with MTL models achieving statistically significant improvement in F1 using the \textit{Gated Double Encoder Model} MTL$_{Gated DEncoder}$ ($p<0.05$, using a paired t-test). This suggests that affective features from the shared encoder benefit the abuse detection task.

\begin{table}[t]
\centering
\begin{subtable}{0.48\textwidth}
\begin{tabular}{|l|l|l|l|}
 \hline
\textbf{Model} & \textbf{P} & \textbf{R} & \textbf{F1} \\ \hline
STL$_{BiLSTM+attn}$ & 77.40 & 73.27 & 74.40  \\ \hline
MTL$_{Hard}$ & 77.21 & 73.30 & 74.51  \\ 
MTL$_{DEncoder}$ & \textbf{77.47} & 73.82 & 74.97 \\ 
MTL$_{Gated DEncoder}$ & 77.46 & \textbf{75.27} & \textbf{76.03}\textsuperscript{\textdagger} \\ \hline
\end{tabular} 
\caption{{\normalsize \textit{Twitter - OffensEval} results.}}
\label{tab:mtl_offenseval_emotion}
\end{subtable}
\bigskip

\begin{subtable}{0.48\textwidth}
\begin{tabular}{|l|l|l|l|} \hline
\textbf{Model} & \textbf{P} & \textbf{R} & \textbf{F1} \\ \hline
STL$_{maxpool+MLP}$ & 79.39 & 78.20 & 78.33  \\ \hline
MTL$_{Hard}$ & 79.34 & 77.61 & 77.90  \\ 
MTL$_{DEncoder}$ & \textbf{80.77} & 78.18 & 79.02   \\ 
MTL$_{Gated DEncoder}$ & 80.12 & \textbf{79.60} & \textbf{79.55}\textsuperscript{\textdagger} \\ \hline
\end{tabular}
\caption{{\normalsize \textit{Twitter - Waseem and Hovy} results.}}
\label{tab:mtl_waseem_emotion}
\end{subtable}

\caption{STL vs. MTL with emotion detection as the auxiliary task. $\dagger$ indicates statistically significant improvement over STL.}
\label{tab:mtl}
\end{table}

\begin{table}[t]
\centering
\resizebox{0.48\textwidth}{!}{
\begin{tabular}{|l|l|l|l|l|}\hline
\textbf{Dataset} & \textbf{Method}  & \textbf{P} & \textbf{R} & \textbf{F1} \\ \hline
\multirow{2}{*}{\textit{OE}} & MTL  & \textbf{77.46} & \textbf{75.27}\textsuperscript{\textdagger}  & \textbf{76.03}\textsuperscript{\textdagger}   \\ \cline{2-5}
 & Transfer & 76.81 & 73.71 & 74.67 \\ \hline
\multirow{2}{*}{\textit{W\&H}} & MTL & 80.12 & \textbf{79.60}\textsuperscript{\textdagger}  & \textbf{79.55}  \\ \cline{2-5} 
 & Transfer & \textbf{81.28} & 77.72 & 79.07 \\ \hline
\end{tabular}}
\caption{MTL vs.~transfer learning performance. \textit{OE} refers to the \textit{OffensEval} dataset and \textit{W\&H} to the \textit{Waseem\&Hovy} dataset. $\dagger$ indicates statistically significant improvements. }
\label{tab:transfer}
\end{table} 

\begin{table*}[!htb]
\begin{tabularx}{\textwidth}{X|l|l|l|l}
\textbf{Sample} & \textbf{STL} & \textbf{MTL} & \textbf{Gold Label} & \textbf{Predicted Emotion} \\ \hline
\textit{\textbf{Shut up} Katie and Nikki... That is all :) \#HASHTAG} & \textit{neither} & \textit{sexism} & \textit{sexism} & \textit{disgust} \\ \hline
\textit{\textit{\_MTN\_} That's the disadvantage of following a religion of uneducated \textbf{morons}, so that you have to rely on Kufir for everything.} & \textit{neither} & \textit{racism} & \textit{racism} & \textit{anger}, \textit{disgust} \\ \hline
\textit{\textit{\_MTN\_} Earthly tyrants want to be \textbf{feared} because for them \textbf{fear} is control and obedience. The writer of the Quran was unsophisticated.} & \textit{neither} & \textit{racism} & \textit{racism} & \textit{fear}, \textit{optimism}\\ \hline
\textit{\textit{\_MTN\_} And does this \textbf{surprise} any of us \textit{\_POLITICIAN\_} SUPPORTERS!!! Not at all... We have heard him accused of everything that can be imagined!!! We still stand BEHIND \textit{\_POLITICIAN\_}!!!} & \textit{Offensive} & \textit{NotOffensive} & \textit{NotOffensive} & None \\ \hline
\textit{\textit{\_MTN\_} I m pretty sure you are not too bad yourself...\textbf{thanks} for a lil bit of \textbf{sweetness} on this brutal world} & \textit{Offensive} & \textit{NotOffensive} & \textit{NotOffensive} & \textit{joy}, \textit{optimism}\\ \hline
\end{tabularx}

\caption{STL vs. MTL: samples from \textit{Twitter - Waseem and Hovy} and \textit{Twitter - OffensEval} datasets, where superior performance of MTL is observed. The `predicted emotion' column contains the emotion labels predicted on the abuse detection data. The name of the politician in the fourth row is masked using the \textit{\_POLITICIAN\_} tag.}
\label{tab:mtl_emotion_improved_examples}
\end{table*}
\normalsize

\paragraph{MTL vs. transfer learning}
Transfer learning is an alternative to MTL that also allows us to transfer knowledge from one task to another. This experiment aims to compare the effectiveness of MTL against transfer learning. We selected the MTL model with the best performance in abuse detection and compared it against an identical model, but trained in a transfer learning setting. In this setup, we first train the model on the emotion detection task until convergence and then proceed by fine-tuning it for the abuse detection task. Table \ref{tab:transfer} presents the comparison between MTL and transfer learning, for which we use the same architecture and hyperparameter configuration as MTL. We observe that MTL outperforms transfer learning and provides statistically significant $(p < 0.05)$ results on both \textit{OffensEval} and \textit{Waseem and Hovy} datasets. 

\section{Discussion} 
\paragraph{Auxiliary  task} Our results show that emotion detection significantly improves abuse detection on both \textit{OffensEval} and \textit{Waseem and Hovy} datasets. Table \ref{tab:mtl_emotion_improved_examples} presents examples of improvements in both datasets achieved by the MTL$_{Gated DEncoder}$ model, over the STL model. In the examples, the highlighted words are emotion evocative words, which are also found in the \textit{SemEval2018 Emotion} dataset. As the emotion detection task encourages the model to learn to predict the emotion labels for the examples that contain these words, the word representations and encoder weights that are learned by the model encompass some affective knowledge. Ultimately, this allows the MTL model to determine the affective nature of the example, which may help it to classify abuse more accurately. It is also interesting to observe that a controversial person or topic may strongly influence the classification of the sample containing it. For example, sentences referring to certain politicians may be classified as \textit{Offensive}, regardless of the context. An example instance of this can be found in Table \ref{tab:mtl_emotion_improved_examples}.\footnote{We mask the name using the \textit{\_POLITICIAN\_} tag.} The MTL model, however, classifies it correctly, which may be attributed to the excessive use of ``\textit{!}'' marks. The latter is one of the most frequently used symbols in the \textit{SemEval2018 Emotion} dataset, and it can encompass many emotions such as \textit{surprise}, \textit{fear}, etc., therefore, not being indicative of a particular type of emotion. Such knowledge can be learned within the shared features of the MTL model. 

\paragraph{MTL vs. transfer learning}
This experiment demonstrates that MTL achieves higher performance than transfer learning in a similar experimental setting. The higher performance may be indicative of a more stable way of transferring knowledge, which leads to better generalization. In the MTL framework, since the shared parameters are updated alternately, each task learns some knowledge that may be mutually beneficial to both related tasks, which leads to a shared representation that encompasses the knowledge of both tasks and hence is more generalized. In contrast, in the case of transfer learning, the primary task fine-tunes the knowledge from the auxiliary task (i.e., in the form of pre-trained parameters) for its task objective and may be forgetting auxiliary task knowledge.

\section{Conclusion}
In this paper, we proposed a new approach to abuse detection, which takes advantage of the affective features to gain auxiliary knowledge through an MTL framework. Our experiments demonstrate that MTL with emotion detection is beneficial for the abuse detection task in the \textit{Twitter} domain. The mutually beneficial relationship that exists between these two tasks opens new research avenues for improvement of abuse detection systems in other domains as well, where emotion would equally play a role. Overall, our results also suggest the superiority of MTL over STL for abuse detection. With this new approach, one can build more complex models introducing new auxiliary tasks for abuse detection. For instance, we expect that abuse detection may also benefit from joint learning with complex semantic tasks, such as figurative language processing and inference. 

\bibliography{acl2020}
\bibliographystyle{acl_natbib}
\appendix
\end{document}